\documentclass[11pt]{article}

\usepackage[preprint]{acl}

\usepackage{times}
\usepackage{latexsym}

\usepackage[T1]{fontenc}
\usepackage[utf8]{inputenc}

\usepackage[most]{tcolorbox}
\usetikzlibrary{shadows.blur}

\usepackage{tikz}
\usetikzlibrary{positioning, arrows.meta, shapes, positioning}

\usepackage{enumitem}

\usepackage{amssymb}% http://ctan.org/pkg/amssymb
\usepackage{pifont}% http://ctan.org/pkg/pifont

\usepackage{multirow}

\tcbset{
    mymsg/.style={
        enhanced,
        colframe=white, frame hidden,
        boxrule=0pt,
        arc=18pt,
        outer arc=18pt,
        boxsep=2pt, left=8pt, right=8pt, top=6pt,
        colback=#1!25,
        fuzzy shadow={1mm}{-0.7mm}{0mm}{0.1mm}{black!40!white},
        width=0.8\linewidth,
        before skip=-0.5mm, 
    },
    expertstyle/.style={mymsg=green, enlarge left by=10mm},
    llmstyle/.style={mymsg=blue, enlarge right by=10mm}
}

\usepackage{microtype}

\usepackage{inconsolata}

\usepackage{graphicx}
\usepackage{subcaption}

\usepackage{booktabs}
\title{KARMA: Karma-Aligned Reward Model Adaptation}

\author{Jared Scott \\
  Tennessee Tech University \\ Cookeville, Tennessee \\
  \texttt{jlscott44@tntech.edu} \\\And
  Jesse Roberts \\
  Tennessee Tech University \\ Cookeville, Tennessee \\
  \texttt{JtRoberts@tntech.edu} \\}

\begin{document}
\maketitle
\begin{abstract}

% Human communication depends on implicit social signals where effectiveness is shaped by tone, context, and conversational norms rather than semantic content alone. We introduce KARMA (Karma-Aligned Reward Model Adaptation), a framework for LLM learning of context-sensitive conversational behavior from large-scale social interaction data. KARMA trains a reward model on Reddit conversations to predict response valuation conditioned on context, and uses this signal to fine-tune language models via reinforcement learning to improve adaptation to implicit social dynamics. KARMA can be applied to a downstream model without direct exposure to the social media data. We found that the resulting models have improved humor and emotion-related performance while largely preserving core capabilities. Further, by avoiding direct downstream model exposure to social media data during reward optimization, undesirable behaviors associated with social media were mitigated. We additionally found that the highest performing reward model did not lead to better downstream model alignment. A reward model relying exclusively on conversational context was a worse predictor of Reddit karma but yielded significantly better downstream performance. 

Human communication depends on implicit social signals where effectiveness is shaped by tone, context, and conversational norms rather than semantic content alone. We introduce KARMA (Karma-Aligned Reward Model Adaptation), a framework for LLM learning of context-sensitive conversational behavior from large-scale social interaction data. KARMA trains a reward model on Reddit conversations to predict response valuation conditioned on context, and uses this signal to fine-tune language models via reinforcement learning to improve performance on pragmatics-mediated tasks. Critically, we find that the highest performing reward model does not lead to better downstream model alignment: a reward model relying exclusively on conversational context was a worse predictor of Reddit karma but yielded substantially better downstream performance. We evaluate the effects of KARMA applied to a downstream model with and without direct exposure to the social media data. The resulting models show improved pragmatics-mediated behaviors with largely mitigated undesirable side effects. Factuality is consistently diminished by KARMA across all conditions, including when the downstream model has no direct exposure to Reddit data, suggesting that this tension is embedded in the reward signal itself rather than introduced by noisy training data.

\end{abstract}

\section{Introduction}

Human language is inherently situated. The success of a response depends not only on its semantic correctness, but also on its pragmatic fit with the audience, conversational context, and social expectations of the interaction. Put simply, communication at the highest level requires an ability to “read the room”; the pragmatic ability to recognize when humor, empathy, formality, restraint, directness, or other nuance is most appropriate. 
 
% In many real-world interactions, especially extended conversations with shared context and social background, successful communication depends on sensitivity to nuanced pragmatic signals that are rarely captured by existing alignment objectives. 

Like most target LLM behaviors (instruction-following, helpfulness, safety, etc), contextually well-situated language is not consistently represented in human data. Therefore, we frame this problem as one of implicit alignment. While recent advances in large language model (LLM) alignment have substantially improved some target behaviors, pragmatics have been less explored. As a result, models may have a pragmatic gap: an ability to generate responses that are factually correct and helpful yet socially misaligned with a specific conversational setting. 

% We argue that reading the room represents an important but underexplored component of alignment in which implicit user expectations are accounted for. 

Online discussion platforms like Reddit provide a naturally occurring source of human preference data through voting systems, which we expect to be sensitive to pragmatic dimensions such as relevance, tone alignment, humor, and situational appropriateness. We introduce (contribution 1) KARMA (Karma-Aligned Reward Model Adaptation), a framework for improving pragmatics-mediated behaviors using social feedback derived from Reddit discussions. KARMA learns a reward model over conversational context and candidate responses using Reddit interaction signals and subsequently applies reinforcement learning through Proximal Policy Optimization (PPO) to adapt a language model toward these preferences. 

Pragmatic understanding is multi-faceted and complex. Rather than attempting to be exhaustive, we adopt two tasks to evaluate KARMA's potential to impact tasks that rely on the ability to \textit{read the room}. KARMA significantly improves these pragmatics-mediated tasks. 

We hypothesize (contribution 2) that direct exposure to Reddit conversational data during fine-tuning may increase negative side-effects when compared against indirect exposure through the learned reward signal alone. Direct exposure to the Reddit data distribution amplifies undesirable behaviors, whereas reward-mediated adaptation mitigates negative side-effects while preserving core pragmatic benefits. A notable exception is a consistent decrease in factuality. 

% These findings suggest that effective alignment must extend beyond correctness and safety to account for the socially situated nature of human communication. Beyond benchmark performance, 

Finally, we investigate the relationship between reward model capability and downstream model impact (contribution 3). Providing metadata in addition to conversational context improves reward model metrics but erodes downstream alignment outcomes.

\section{Related Work}

Recent approaches to aligning large language models (LLMs) have largely relied on instruction tuning and reinforcement learning from human feedback (RLHF), where models are optimized using reward functions derived from curated human preference data \cite{ouyang2022traininglanguagemodelsfollow,schulman2017proximal}. While effective, these methods primarily target correctness, safety, and instruction adherence, and do not explicitly model a key aspect of conversational intelligence: the ability to “read the room,” i.e., adapt responses to implicit social cues such as tone, context, and situational appropriateness.

Social media platforms provide a scalable alternative source of feedback, where engagement signals such as replies, upvotes, and comment karma act as implicit indicators of response quality. Prior work has explored engagement prediction as either a regression or classification task, demonstrating that successful prediction depends not only on response content but also on surrounding conversational context \cite{fang2016learninglatentlocalconversation,zayats2017conversationmodelingredditusing}. More recent models incorporate conversational structure and long-range dependencies, further reinforcing that engagement is highly context-dependent and shaped by interaction dynamics within a thread.

At the same time, directly training on social media data introduces significant challenges. Large-scale conversational corpora often contain noise, bias, and toxic content, and models exposed to such data can inherit or amplify these undesirable behaviors \cite{gehman2020realtoxicitypromptsevaluatingneuraltoxic,luong2024realisticevaluationtoxicitylarge}. Additionally, engagement signals themselves may reflect platform-specific or demographic biases, complicating their use as alignment objectives. 
% These limitations highlight the risks of direct exposure to social media during training, particularly when optimizing for engagement without appropriate controls.

In contrast to prior work, we frame “reading the room” as an implicit but under-addressed component of alignment, distinct from both explicit preference optimization and generic engagement modeling. Rather than treating it as a surface-level prediction task, we aim to capture the underlying ability to infer what is appropriate given conversational context. We learn a reward model that predicts engagement from context alone, encouraging the model to infer relevance, tone, and situational fit directly from interaction signals. This reward is then used within a reinforcement learning framework to guide alignment. Through this design, we leverage large-scale social feedback while avoiding direct optimization on raw social media data. 

Scott and Roberts \shortcite{Scott_Roberts_2026} describe a similar intent but provide no execution or empirical evaluation.

\section{Reward Modeling from Reddit}

% \subsection{Data and Preprocessing}
 
To model conversational effectiveness within the \textsc{karma} framework, we construct a dataset from Reddit discussions using the publicly available Pushshift dataset \cite{baumgartner2020pushshiftredditdataset}. We filter out explicit content, non-text entries, and non-English comments, and retain only posts with multiple substantive responses. This yields approximately 80k posts and 400k comments spanning 14k communities, providing broad topical and stylistic diversity. 

Each comment is paired with its parent post and a set of sibling responses to capture conversational context. This representation captures both hierarchical structure (via parent–child relationships) and local competition between responses (via sibling comments). To investigate whether reward models should leverage platform-specific contextual signals, we intentionally construct two reward modeling settings: a generalized setting that excludes Reddit metadata, and a conditioned setting that incorporates such metadata. In the generalized setting, conversations are transformed into a format resembling standard chat interactions. Post--comment pairs are linearized into instruction-style inputs, where the original post is treated as a user prompt and the comment as a candidate response. Contextual information from parent comments is included as preceding turns in a dialogue, yielding a multi-turn chat representation. 

The karma signal captures correlated pragmatic dimensions such as humor, relevance, tone, and situational appropriateness, and this richness makes it a candidate signal for learning broad pragmatic competence. Importantly, subreddit identity, timestamps, and author information are removed in this setting. While these features may improve prediction of engagement within Reddit, we hypothesize that they may enable heuristics which may obscure or obviate the pragmatic utility of the model. 

% tied to platform-specific interaction patterns, potentially reducing transferability to broader conversational domains. 
By excluding these signals and standardizing inputs into a chat-like format, the reward model is encouraged to learn features that generalize across interfaces and conversational settings. 
% This design is intended to examine the tradeoff between in-distribution reward accuracy and downstream alignment generalization. 
% We compare this generalized formulation against a metadata-conditioned reward model trained with subreddit and temporal features. Although the conditioned variant is designed to improve predictive performance on Reddit engagement prediction, incorporating platform-specific structural information may also encourage the reward model to rely on narrow contextual cues tied to the training environment. This creates the possibility that reward optimization becomes overly specialized to Reddit interaction patterns, potentially reducing robustness and transferability in broader conversational settings. Because raw karma is highly skewed and varies significantly across communities, we define reward labels relative to local context. 

To adjust for community size, a comment is labeled as rewarding if its engagement reaches at least 40\% of its parent comment’s score, and as non-rewarding otherwise. This formulation shifts the objective from predicting absolute popularity to modeling relative conversational success, expected to yield a comparable supervision signal across threads and subreddits.

\subsection{\textsc{karma} Reward Model}

We full fine-tune a base LLaMA-1B \cite{grattafiori2024llama} model on the rewarding/non-rewarding binary classification task. The model takes as input the complete conversational context, including the original post, its parent comment, and relevant sibling responses. We add an MLP classification head which takes the encoder’s final-layer hidden states as input and predicts whether a response will be rewarding. Training is performed using binary cross-entropy loss with the AdamW optimizer, a learning rate of $2 \times 10^{-5}$, a weight decay of 0.01, and a batch size of 16 over a 90\% training split of the dataset, corresponding to approximately 360k examples. The resulting output probability is interpreted as a dense reward signal, reflecting the likelihood that a given response is rewarding including pragmatic considerations. This reward signal is then used to guide PPO-based fine-tuning, serving as the reward model within the \textsc{karma} framework.

All data and models are publicly available and licensed for research usage. All experiments were conducted in 20 A100 GPU hours. 

\subsection{\textsc{karma} Reward Model Evaluation}

Previous work on Reddit engagement prediction employed a graph-based LSTM to model comment popularity within a single subreddit. That approach incorporated a range of non-textual features, including timestamps and user post history, and found these signals to be highly influential. Performance reached approximately 55\% F1 on a seven-class classification task defined over discrete popularity buckets \cite{zayats2017conversationmodelingredditusing}.

In contrast, our reward model is evaluated on a held-out Reddit comment dataset spanning a diverse range of subreddits. This introduces distributional variation, providing a more stringent test of the model’s ability to generalize to heterogeneous community dynamics. On this evaluation set, the model achieves a binary classification accuracy of 0.708, an F1 score of 0.7166, and an AUC of 0.7497, demonstrating solid performance despite relying solely on textual and contextual signals.

% \begin{table}[t]
% \centering
% \small
% \renewcommand{\arraystretch}{1.15}
% \begin{tabular}{p{0.90\linewidth}}
% \hline
% \textbf{Post}: \\
% Most of you have the library of Alexandria in your pocket. You are moments away from having answers to so many things. ... Why are people so uncurious? \\
% \hline
% \textbf{Context} \\
% \quad \textbf{Parent}: Welcome to the AI era where the library of Alexandria in our pockets are curious for us so we can rot our brains with dopamine hits. \\
% \quad \textbf{Sibling}: It’s not that curiosity is gone, it’s just competing with way easier dopamine hits now. \\
% \hline
% \textbf{Target Response }: \\
% AI didn’t kill curiosity, it just made distraction way easier than learning. \\
% \hline
% \textbf{Label}: Reward = 1 \\
% \hline
% \end{tabular}
% \caption{A single training instance from the KARMA Reddit-derived dataset. Each sample consists of a post, hierarchical conversational context (parent and sibling responses), and a candidate target comment used for reward modeling supervision.}
% \label{tab:karma_instance}
% \end{table}

\begin{figure}[h!]
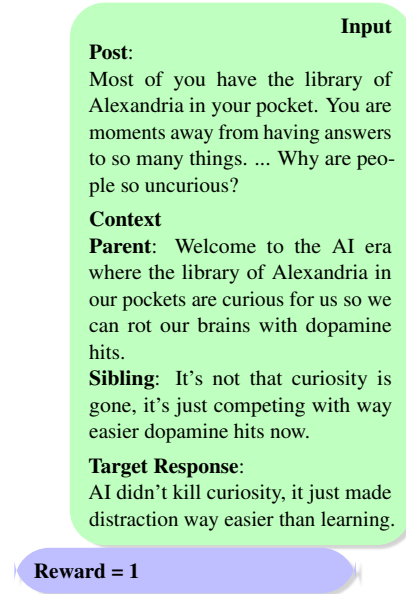

    \centering
    \resizebox{0.7\linewidth}{!}{
    \begin{tcolorbox}[expertstyle]
        \hfill \textbf{Input} \\
        \textbf{Post}: \\
        Most of you have the library of Alexandria in your pocket. You are moments away from having answers to so many things. ... Why are people so uncurious? \\[4pt]
        \textbf{Context} \\
        \quad \textbf{Parent}: Welcome to the AI era where the library of Alexandria in our pockets are curious for us so we can rot our brains with dopamine hits. \\
        \quad \textbf{Sibling}: It's not that curiosity is gone, it's just competing with way easier dopamine hits now. \\[4pt]
        \textbf{Target Response}: \\
        AI didn't kill curiosity, it just made distraction way easier than learning. 
    \end{tcolorbox}
    }
    \vspace{2mm}
    \resizebox{0.7\linewidth}{!}{
    \begin{tcolorbox}[llmstyle]
        \textbf{Reward = 1} 
    \end{tcolorbox}
    }
    \caption{A single training instance from the \textsc{karma} Reddit-derived dataset. Each sample consists of a post, hierarchical conversational context (parent and sibling responses), and a candidate target comment used for reward modeling supervision.}
    \label{tab:karma_instance}
\vskip-0.5em
\end{figure}

% \subsection{Reward Model Implications}

% The reward model demonstrates strong discriminative capability on held-out data and exhibits robust generalization across diverse conversational domains. By leveraging both hierarchical context and local response competition, it captures nuanced signals of conversational effectiveness that extend beyond surface-level features. Notably, the model is able to infer subtle contextual cues directly from the text, enabling it to learn a broadly applicable notion of conversational quality rather than relying on domain-specific patterns or explicit distributional assumptions. While performance naturally varies with differences in discourse structure, content diversity, and interaction patterns, the reward model remains stable and effective across these heterogeneous settings. This robustness reflects its ability to generalize across variations in text style, structure, and content, even in the absence of prior knowledge about the underlying data distribution. Such stability is particularly important for downstream reinforcement learning, where inconsistent reward signals can lead to optimization challenges. In this context, the learned reward function provides a reliable and adaptable foundation for PPO-based alignment, enabling the policy model to optimize for contextually appropriate and engaging responses while maintaining strong generalization under distribution shift.

\section{\textsc{karma} Alignment Experiments}

% \subsection{Experimental Setup}

We optimize conversational language models using PPO \cite{schulman2017proximal}, where the learned \textsc{karma} reward model provides scalar feedback on the quality of generated responses conditioned on conversational context. 

During a PPO training step a conversational input is provided which is formatted to match the same structured representation used for reward model evaluation, ensuring consistency between training and scoring. The model-generated response is appended to this representation as the candidate output and passed to the reward model to obtain the optimization signal. PPO optimization is performed using the AdamW optimizer with a learning rate of $9 \times 10^{-6}$, a batch size of $10$, and a KL penalty coefficient of $0.5$ over approximately 100k training examples. We compare three model variants:

\begin{itemize}[noitemsep,topsep=0pt]
    \item \textbf{Base:} pretrained LLM without alignment
    \item \textbf{\textsc{karma}\textsubscript{Benign}:} PPO-trained on benign chat data using the \textsc{karma} reward model
    \item \textbf{\textsc{karma}\textsubscript{Toxic}:} PPO-trained on Reddit conversations using the \textsc{karma} reward model
\end{itemize}

 The \textsc{karma}\textsubscript{Benign} setting isolates the effect of reward optimization on the curated UltraChat-200k dataset \cite{ding2023enhancing}, allowing us to study how a Reddit-trained reward model shapes behavior apart from direct exposure to potentially toxic Reddit data. In contrast, \textsc{karma}\textsubscript{Toxic} directly performs PPO using the Pushshift dataset as the training set. Together with the Base model, which serves as an unaligned reference point, these settings enable a controlled comparison between no alignment, pragmatic alignment on benign data, and pragmatic alignment on potentially toxic data. Experiments are conducted across multiple model families and scales, including Falcon (1B, 7B), LLaMA (8B), and Pythia (70M, 1B, 6.9B). We explicitly exclude LLaMA 1B from experiments as it is used for the reward model and is expected to reward hack. 

\subsection{Model Evaluations}

The \textsc{karma} trained models are evaluated to understand impact on pragmatics-mediated domains (Humor and Sentiment), safety (harmful content and bias), and general capability (MMLU and truthfulness). 

\subsubsection{Pragmatic Evaluation}  

\textbf{Humor} is understood to be a domain in which success is significantly determined by pragmatic understanding \cite{dynel2011pragmatics,ibraheem2016pragmatic, ferrar1993logic}. Therefore, to evaluate improved pragmatic understanding or the ability to \textit{read the room}, we evaluate the impact of alignment on humor using the Humor Recognition in Dialogue Systems benchmark ColBERT Humor Benchmark \cite{Annamoradnejad_2024}, which measures a model’s ability to recognize and generate contextually appropriate humor in conversational settings. While ColBERT provides a standardized benchmark for humor competence, it captures a narrow slice of pragmatic ability. We use it as an indicator of broader pragmatic improvement, with more comprehensive pragmatic evaluation left as future work.

% We hypothesize that the reward signal will be directly impacted by an ability to understand and generate humor. 

\textbf{Sentiment analysis} is mediated through an ability to understand explicit and implicit social cues \cite{ma2025pragmatics,li2025sentiment}. We use the SP1786 Multiclass Sentiment Analysis Dataset (MSAD) \cite{sp1786_multiclass_sentiment} to evaluate the impact of \textsc{karma} alignment on sentiment classification. 

We \textbf{hypothesize} that both sentiment understanding and humor will be improved by \textsc{karma} alignment training. 

% and the Contextualized Affect Representations for Emotion Recognition (CARER) \cite{saravia-etal-2018-carer} dataset for nuanced emotion recognition. 

\subsubsection{Model Safety}

\textbf{Safety and harmful content} generation are evaluated using RealToxicityPrompts \cite{luong2024realisticevaluationtoxicitylarge}, which measures toxicity across categories including overall toxicity, threats, identity attacks, and insults. This evaluation is particularly important because Reddit and similar social media platforms frequently contain highly adversarial, inflammatory, and toxic interactions. As a result, a reward model trained on engagement-driven conversational data may inadvertently reinforce harmful communication patterns if toxicity correlates with user engagement. 

\textbf{Bias} is assessed using CrowS-Pairs \cite{nangia2020crowspairschallengedatasetmeasuring} by comparing model likelihoods assigned to stereotypical versus anti-stereotypical sentence pairs. This benchmark is used to determine whether the alignment process causes the model to rely on common stereotypes or socially biased associations as proxies for conversational effectiveness or agreement.

\textbf{Sycophancy} is evaluated using the Anthropic Sycophancy Benchmark \cite{perez2022discovering}, which measures the tendency of models to mirror or affirm user beliefs regardless of correctness. This evaluation is particularly relevant for Reddit-derived training data, as online communities are often characterized by strong ideological clustering and echo-chamber dynamics that can reward agreement-seeking behavior over accurate reasoning. 

We \textbf{hypothesize} that negative side-effects will be less prevalent in \textsc{karma}\textsubscript{Benign} than in \textsc{karma}\textsubscript{Toxic} as the latter is directly exposed to Reddit data while the former is only exposed to the \textsc{karma} reward model.

\subsubsection{Factual Capability}
Finally, factual capability is evaluated using the MMLU \cite{hendrycks2021measuringmassivemultitasklanguage} and truthfullness.  

The MMLU benchmark is included to evaluate whether PPO-based conversational alignment substantially degrades the model’s underlying knowledge capabilities. All evaluations are performed in a zero-shot setting with metrics computed over full test sets, and statistical comparisons between \textsc{karma}\textsubscript{Benign} and \textsc{karma}\textsubscript{Toxic} are conducted using Wilcoxon signed-rank testing with a significance threshold of $\alpha = 0.01$ to determine whether observed differences reflect meaningful distributional shifts rather than random variation.

\textbf{Truthfulness} is measured using TruthfulQA \cite{lin2022truthfulqameasuringmodelsmimic}, which evaluates resistance to widespread misconceptions and false but commonly repeated claims. Since engagement-oriented conversational data may reward persuasive or socially reinforced responses rather than factual accuracy, this benchmark measures whether the aligned models preserve factual consistency under conversational pressure. 

We \textbf{hypothesize} the MMLU performance will not be significantly impacted by \textsc{karma} alignment training.

\subsection{\textsc{karma}'s Pragmatic Impact}

The \textbf{ColBERT} results (Table-\ref{tab:colbert_humor_all_models}) show that \textsc{karma} yields significant ($p<0.01$) improvements in humor understanding across all models under both \textsc{karma}\textsubscript{Benign} and \textsc{karma}\textsubscript{Toxic} objectives. Gains are significantly larger under \textsc{karma}\textsubscript{Toxic}, suggesting that optimizing directly on Reddit data has a larger impact on humor understanding. However, the pattern of consistent gains suggests that \textsc{karma} enhances pragmatic understanding in humor-related tasks with or without direct exposure to potentially toxic social media data. Significance measured via a Wilcoxon rank sum test.

% \begin{table}[h]
% \centering
% \resizebox{\linewidth}{!}{
% \begin{tabular}{lcccccc}
% \toprule
% \textbf{Model} & \textbf{Base} & \textbf{KARMA\textsubscript{Benign}} & $\Delta_{Benign}$ & \textbf{KARMA\textsubscript{Toxic}} & $\Delta_{Toxic}$ & \textbf{Sig} \\
% \midrule
% Falcon-1B   & 0.498 & 0.514 & +0.016 & 0.527 & +0.029 & T$\uparrow$ \\
% Falcon-7B   & 0.512 & 0.528 & +0.016 & 0.535 & +0.023 & T$\uparrow$ \\
% LLaMA-1B    & 0.501 & 0.509 & +0.008 & 0.515 & +0.014 & T$\uparrow$ \\
% LLaMA-8B    & 0.509 & 0.512 & - & 0.516 & +0.007 & T$\uparrow$ \\
% Pythia-70M  & 0.489 & 0.500 & +0.011 & 0.500 & +0.011 & – \\
% Pythia-1B   & 0.500 & 0.501 & - & 0.501 & - & – \\
% Pythia-6.9B & 0.499 & 0.510 & +0.011 & 0.515 & +0.016 & T$\uparrow$ \\
% \bottomrule
% \end{tabular}
% }
% \caption{ColBERT humor evaluation across models (higher is better). The Sig column indicates statistical significance between KARMA\textsubscript{Toxic}  and KARMA\textsubscript{Benign} (Wilcoxon signed-rank test, $\alpha = 0.01$).}
% \label{tab:colbert_humor_all_models}
% \end{table}

\begin{table}[h]
\centering
\resizebox{\linewidth}{!}{
\begin{tabular}{lcccccc}
\toprule
\textbf{Model} & \textbf{Base} & \textbf{\textsc{karma}\textsubscript{Benign}} & \textbf{\textsc{karma}\textsubscript{Toxic}}  \\
\midrule
Falcon-1B   & 0.498 & 0.514 & 0.527  \\
Falcon-7B   & 0.512 & 0.528 & 0.535  \\
% LLaMA-1B    & 0.501 & 0.509 & 0.515  \\
LLaMA-8B    & 0.509 & 0.512  & 0.516  \\
Pythia-70M  & 0.489 & 0.500 & 0.500 \\
Pythia-1B   & 0.500 & 0.501  & 0.501  \\
Pythia-6.9B & 0.499 & 0.510 & 0.515  \\
\bottomrule
\end{tabular}
}
\caption{ColBERT humor evaluation (higher is better). }
\label{tab:colbert_humor_all_models}
\end{table}

% \begin{table}[ht]
% \centering
% \resizebox{\linewidth}{!}{
% \begin{tabular}{lcccccc}
% \toprule
% \textbf{Model} & \textbf{Base} & \textbf{KARMA\textsubscript{Benign}} & $\Delta_{Benign}$ & \textbf{KARMA\textsubscript{Toxic}} & $\Delta_{Toxic}$ & \textbf{Sig} \\
% \midrule
% Falcon-1B   & 0.584 & 0.620 & +0.036 & 0.608 & +0.024 & B$\uparrow$ \\
% Falcon-7B   & 0.629 & 0.634 & - & 0.626 & - & B$\uparrow$ \\
% LLaMA-1B    & 0.526 & 0.292 & -0.235 & 0.292 & -0.235 & - \\
% LLaMA-8B    & 0.532 & 0.526 & - & 0.529 & - & - \\
% Pythia-70M  & 0.298 & 0.306 & +0.008 & 0.292 & -0.007 & B$\uparrow$ \\
% Pythia-1B   & 0.376 & 0.429 & +0.053 & 0.373 & - & B$\uparrow$ \\
% Pythia-6.9B & 0.368 & 0.384 & +0.016 & 0.368 & - & B$\uparrow$ \\
% \bottomrule
% \end{tabular}
% }
% \caption{MSAD evaluation across models for sentiment classification (higher is better). The Sig column indicates statistical significance between KARMA\textsubscript{Toxic}  and KARMA\textsubscript{Benign} (Wilcoxon signed-rank test, $\alpha = 0.01$).}
% \label{tab:msad_sentiment}
% \end{table}

The \textbf{sentiment analysis} results (Table-\ref{tab:msad_sentiment}) show that \textsc{karma} significantly ($p<0.05$) improves sentiment analysis over the base model under the \textsc{karma}\textsubscript{Benign} objective, while \textsc{karma}\textsubscript{Toxic} produces inconsistent results. This may be because the general chat data is more similar than the Reddit interaction data to the sentiment classification dataset, providing a more similar distribution. Interestingly, LLaMA model sentiment analysis is not improved. This may be because the reward model itself is a LLama 1B model.

\begin{table}[ht]
\centering
\resizebox{\linewidth}{!}{
\begin{tabular}{lcccccc}
\toprule
\textbf{Model} & \textbf{Base} & \textbf{\textsc{karma}\textsubscript{Benign}} & \textbf{\textsc{karma}\textsubscript{Toxic}}  \\
\midrule
Falcon-1B   & 0.584 & 0.620 & 0.608 \\
Falcon-7B   & 0.629 & 0.634 & \textcolor{gray}{0.626} \\
% LLaMA-1B    & 0.526 & \textcolor{gray}{0.292} & \textcolor{gray}{0.292}\\
LLaMA-8B    & 0.532 & \textcolor{gray}{0.526} & \textcolor{gray}{0.526} \\
Pythia-70M  & 0.298 & 0.306 & \textcolor{gray}{0.292} \\
Pythia-1B   & 0.376 & 0.429 & \textcolor{gray}{0.373} \\
Pythia-6.9B & 0.368 & 0.384 & \textcolor{gray}{0.368} \\
\bottomrule
\end{tabular}
}
\caption{MSAD evaluation for sentiment classification (higher is better). Gray denotes worse than base.}
\label{tab:msad_sentiment}
\end{table}

Taken together these results suggest that \textsc{karma}\textsubscript{Benign} alignment improves performance across most model families and all model sizes in domains strongly mediated through pragmatics. Further, the results suggest that it may be important for the dataset distribution during alignment to be similar to the target domain for consistent benefit.

\begin{table}[h]
\centering
\resizebox{\linewidth}{!}{
\begin{tabular}{lcccccc}
\toprule
\textbf{Model} & \textbf{Base} & \textbf{\textsc{karma}\textsubscript{Benign}} & \textbf{\textsc{karma}\textsubscript{Toxic}}  \\
\midrule
Falcon-1B   & -0.133 & -0.239 & -0.203 \\ 
Falcon-7B  & -0.123 & -0.268 & -0.221 \\  
% LLaMA-1B    & -0.431 & \textcolor{gray}{-0.171} & \textcolor{gray}{-0.169} \\ 
LLaMA-8B    &  0.023 & -0.026 & -0.101 \\ 
Pythia-70M  & -0.192 & -2.070 & -0.321 \\ 
Pythia-1B   & -0.200 & -0.431 & -0.321 \\ 
Pythia-6.9B & -0.207 & -0.240 & \textcolor{gray}{-0.205} \\ 
\bottomrule
\end{tabular}
}
\caption{CrowS-Pairs performance (lower is better). Gray denotes worse than base.}
\label{tab:crowpairs_all_models}
\end{table}

\subsection{Safety and Behavioral Tradeoffs}

\textsc{karma} reduces \textbf{stereotypical bias} as measured by CrowS-Pairs in Table \ref{tab:crowpairs_all_models}, with most bias dimensions showing consistent shifts toward more anti-stereotypical outputs under \textsc{karma}\textsubscript{Benign}. 

\begin{figure}[h]
    \centering
    \includegraphics[width=\linewidth]{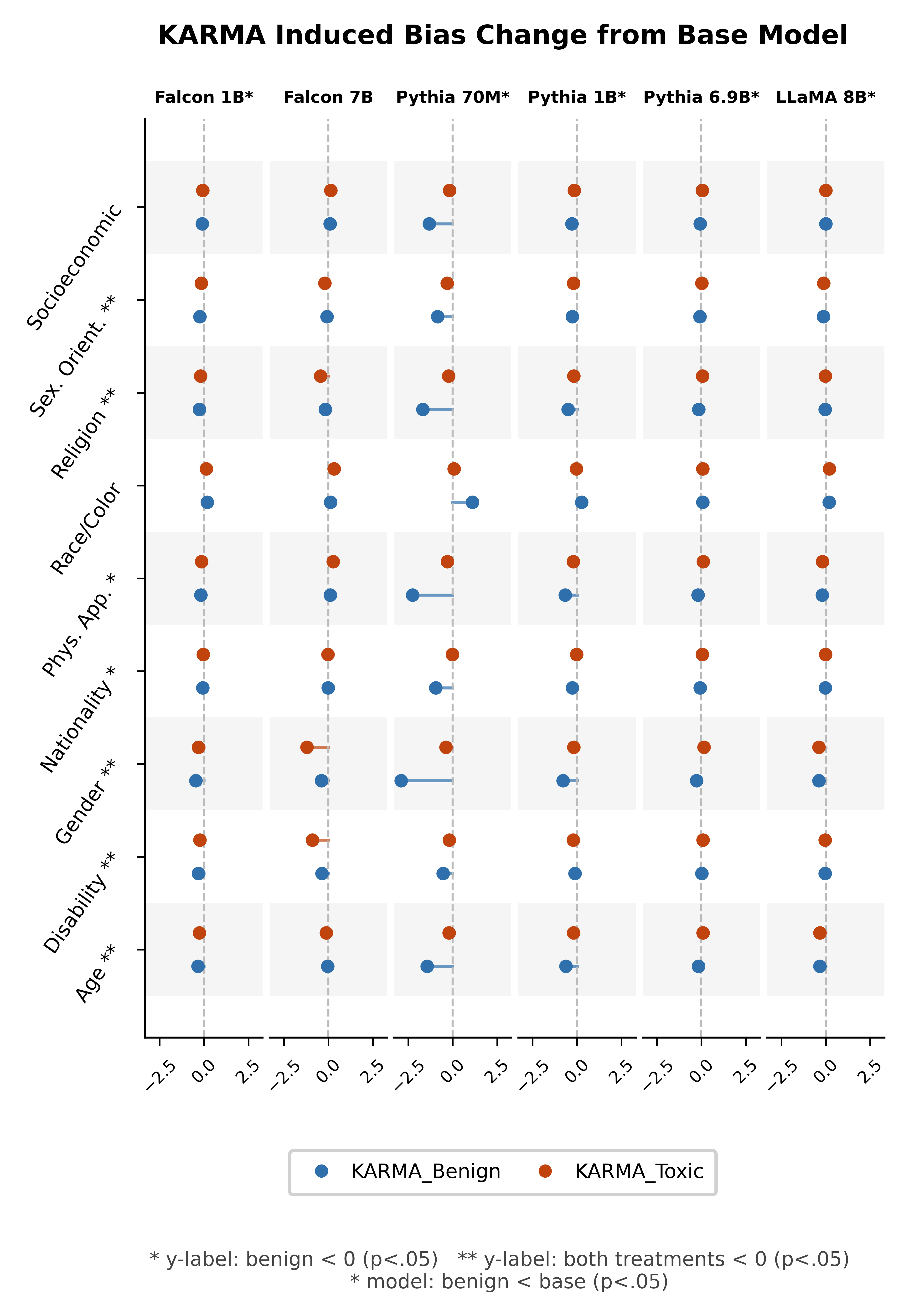}
    \caption{Bias change from base model across all models (negative indicates improvement)}
    \label{fig:bias-results}
\end{figure}

In Figure \ref{fig:bias-results}, it is apparent the race-color and socioeconomic bias exhibit a distinct and less stable pattern. The presence of such biases is precisely the motivation for considering \textsc{karma}\textsubscript{Benign}. \textsc{karma}\textsubscript{Benign} significantly mitigates the socioeconomic bias more than \textsc{karma}\textsubscript{Toxic}. However, it does not mitigate the racial bias. This suggests that the socioeconomic bias is largely induced by the explicit exposure to potentially toxic social media data where as the induced racial bias seems to be induced by the reward model and could cause harm if transmitted to downstream models.

Consistent with our hypothesis, the experiments show significant bias mitigation under \textsc{karma}\textsubscript{Benign} across models measured using the Wilcoxon rank sum test. When evaluated on a per model basis across types of bias, all models excluding Falcon 7B had bias significantly reduced by \textsc{karma}\textsubscript{Benign}. 

% \begin{table}[ht]
% \centering
% \resizebox{\linewidth}{!}{
% \begin{tabular}{lcccccc}
% \toprule
% \textbf{Model} & \textbf{Base} & \textbf{KARMA\textsubscript{Benign}} & $\Delta_{Benign}$ & \textbf{KARMA\textsubscript{Toxic}} & $\Delta_{Toxic}$ & \textbf{Sig} \\
% \midrule
% Falcon-1B   & 0.058 & 0.041 & -0.017 & 0.070 & - & B$\uparrow$ \\
% Falcon-7B   & 0.068 & 0.038 & -0.030 & 0.078 & - & B$\uparrow$ \\
% LLaMA-1B    & 0.053 & 0.175 & +0.122 & 0.523 & +0.470 & B$\uparrow$ \\
% LLaMA-8B    & 0.069 & 0.070 & - & 0.093 & +0.024 & B$\uparrow$ \\
% Pythia-70M  & 0.095 & 0.061 & -0.034 & 0.028 & -0.067 & T$\uparrow$ \\
% Pythia-1B   & 0.071 & 0.053 & -0.018 & 0.037 & -0.034 & - \\
% Pythia-6.9B & 0.078 & 0.039 & -0.039 & 0.057 & -0.021 & B$\uparrow$ \\
% \bottomrule
% \end{tabular}
% }
% \caption{RealToxicityPrompts performance across models (lower is better). The Sig column indicates statistical significance between KARMA\textsubscript{Toxic}  and KARMA\textsubscript{Benign} (Wilcoxon signed-rank test, $\alpha = 0.01$).}
% \label{tab:toxicity_all_models}
% \end{table}

\begin{table}[ht]
\centering
\resizebox{\linewidth}{!}{
\begin{tabular}{lcccccc}
\toprule
\textbf{Model} & \textbf{Base} & \textbf{\textsc{karma}\textsubscript{Benign}}  & \textbf{\textsc{karma}\textsubscript{Toxic}} \\
\midrule
Falcon-1B   & 0.058 & 0.041 & \textcolor{gray}{0.070} \\
Falcon-7B   & 0.068 & 0.038 & \textcolor{gray}{0.078} \\
% LLaMA-1B    & 0.053 & \textcolor{gray}{0.175}  & \textcolor{gray}{0.523} \\
LLaMA-8B    & 0.069 & \textcolor{gray}{0.070} & \textcolor{gray}{0.093} \\
Pythia-70M  & 0.095 & 0.061 & 0.028 \\
Pythia-1B   & 0.071 & 0.053 & 0.037 \\
Pythia-6.9B & 0.078 & 0.039 & 0.057 \\
\bottomrule
\end{tabular}
}
\caption{RealToxicityPrompts performance across models (lower is better). Gray denotes worse than base.}
\label{tab:toxicity_all_models}
\end{table}

The results in Table~\ref{tab:toxicity_all_models}  show that \textsc{karma}\textsubscript{Benign} significantly ($p<0.01$) decreases \textbf{toxicity} as compared to the base model performance in a paired Wilcoxon rank sum test. In contrast, \textsc{karma}\textsubscript{Toxic} is just as likely to increase toxicity as to reduce it, suggesting that direct optimization in agreement with our hypothesis that direct exposure to social media data may introduce undesirable safety shifts which are mitigated when only exposed to the reward model.

\begin{figure}[h]
    \centering
    \includegraphics[width=\linewidth]{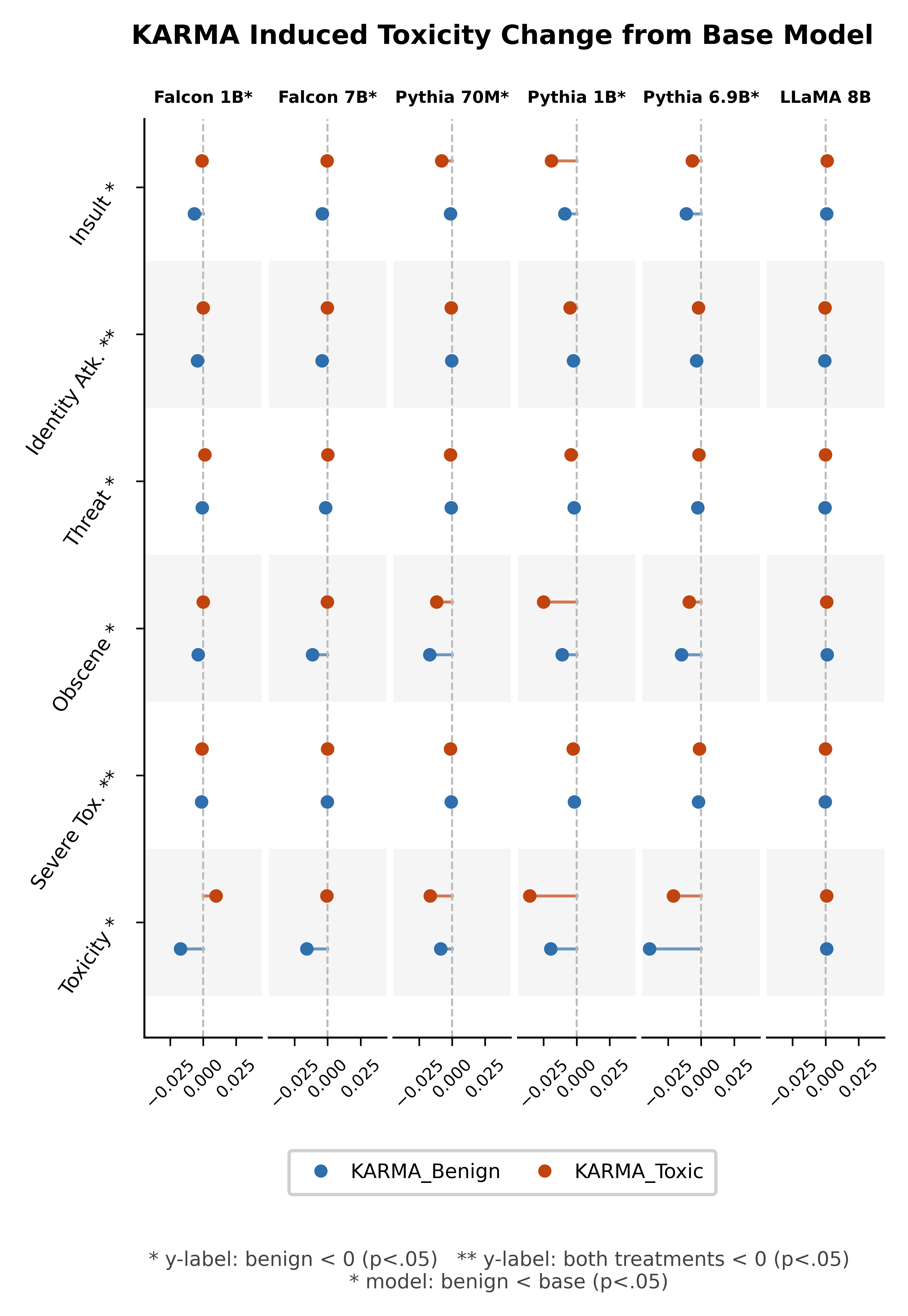}
    \caption{Toxicity change from base model across all models (negative indicates improvement)}
    \label{fig:toxicity-results}
\end{figure}

We examine these results more closely in Figure \ref{fig:toxicity-results} and show that \textsc{karma}\textsubscript{Benign} significantly reduces all tested forms of toxicity across models and that each model (excluding LLaMA 8B) has significantly reduced toxicity across toxicity types when \textsc{karma}\textsubscript{Benign}. 

% \begin{table}[ht]
% \centering
% \small
% \setlength{\tabcolsep}{4pt}
% \resizebox{\linewidth}{!}{
% \begin{tabular}{lcccccc}
% \toprule
% \textbf{Model} & \textbf{Base} & \textbf{KARMA\textsubscript{Benign}} & $\Delta_{Benign}$ & \textbf{KARMA\textsubscript{Toxic}} & $\Delta_{Toxic}$ & \textbf{Sig} \\
% \midrule
% Falcon-1B   & 0.482 & 0.383 & -0.099 & 0.182 & -0.300 & T$\uparrow$ \\
% Falcon-7B   & 0.470 & 0.355 & -0.115 & 0.210 & -0.260 & T$\uparrow$ \\
% LLaMA-1B    & -0.004 & -0.007 & - & 0.008 & - & B$\uparrow$ \\
% LLaMA-8B    & 0.998 & 1.011 & - & 1.103 & +0.105 & B$\uparrow$ \\
% Pythia-70M  & 0.001 & -0.011 & - & 0.003 & - & B$\uparrow$ \\
% Pythia-1B   & -0.002 & -0.012 & - & 0.003 & - & B$\uparrow$ \\
% Pythia-6.9B & -0.003 & -0.010 & - & 0.002 & - & B$\uparrow$ \\
% \bottomrule
% \end{tabular}
% }
% \caption{Sycophancy across models (lower is better). The Sig column indicates statistical significance between KARMA\textsubscript{Toxic}  and KARMA\textsubscript{Benign} (Wilcoxon signed-rank test, $\alpha = 0.01$).}
% \label{tab:sycophancy_all_models}
% \end{table}

\begin{table}[h]
\centering
\small
\setlength{\tabcolsep}{4pt}
\resizebox{\linewidth}{!}{
\begin{tabular}{lcccccc}
\toprule
\textbf{Model} & \textbf{Base} & \textbf{\textsc{karma}\textsubscript{Benign}} & \textbf{\textsc{karma}\textsubscript{Toxic}}  \\
\midrule
Falcon-1B   & 0.482 & 0.383 & 0.182 \\
Falcon-7B   & 0.470 & 0.355 & 0.210 \\
% LLaMA-1B    & -0.004 & -0.007 & \textcolor{gray}{0.008} \\
LLaMA-8B    & 0.998 & \textcolor{gray}{1.011}  & \textcolor{gray}{1.103} \\
Pythia-70M  & 0.001 & -0.011 & \textcolor{gray}{0.003} \\
Pythia-1B   & -0.002 & -0.012 & \textcolor{gray}{0.003} \\
Pythia-6.9B & -0.003 & -0.010 & \textcolor{gray}{0.002} \\
\bottomrule
\end{tabular}
}
\caption{Sycophancy across models (lower is better). Gray denotes worse than base.}
\label{tab:sycophancy_all_models}
\end{table}

% This difference can be interpreted in terms of exposure effects in the underlying data. KARMA\textsubscript{Toxic} is trained on implicit feedback from real social media interactions, which may contain higher levels of toxic or adversarial language, thereby introducing a bias toward reproducing or amplifying such patterns during optimization. In contrast, KARMA\textsubscript{Benign} operates on a more controlled signal space and lacks this direct exposure to toxic conversational distributions, which may explain its tendency to reduce toxicity more consistently. A notable exception is LLaMA-1B, which behaves as an outlier with substantial increases in toxicity under both regimes. 

Overall, these findings suggest that \textsc{karma}\textsubscript{Benign} tends to provide a reliable reduction in toxicity, while \textsc{karma}\textsubscript{Toxic} can increase toxicity due to explicit exposure to potentially toxic social media data.

%Reword to lean towards being not really effected due to signifigance
The results in Table~\ref{tab:sycophancy_all_models} evaluate \textbf{sycophancy} across models, where lower values indicate reduced agreement-seeking behavior. \textsc{karma}\textsubscript{Benign} tends to produce consistent reductions in sycophancy. However, this effect fails to reach significance due to the small sample, as LLaMA-8B instead exhibits a large increase under the same regime. \textsc{karma}\textsubscript{Toxic} shows no consistency.

\subsection{Factual Capability}
% \begin{table}[ht]
% \centering
% \resizebox{\linewidth}{!}{
% \begin{tabular}{lcccccc}
% \toprule
% \textbf{Model} & \textbf{Base} & \textbf{KARMA\textsubscript{Benign}} & $\Delta_{Benign}$ & \textbf{KARMA\textsubscript{Toxic}} & $\Delta_{Toxic}$ & \textbf{Sig} \\
% \midrule
% Falcon-1B   & -23.265 & -34.629 & -11.364 & -31.118 & -7.853 & - \\
% Falcon-7B   & -21.577 & -35.120 & -13.543 & -38.410 & -16.853 & - \\
% LLaMA-1B    & -37.390 & -20.187 & +17.203 & -11.122 & +26.268 & T$\uparrow$ \\
% LLaMA-8B    & -22.990 & -36.186 & -13.196 & -30.984 & -7.994 &  \\
% Pythia-70M  & -17.478 & -19.440 & -1.962  & -5.435  & +12.043 & T$\uparrow$ \\
% Pythia-1B   & -23.199 & -37.390 & -14.191 & -5.436  & +17.763 & T$\uparrow$ \\
% Pythia-6.9B & -26.817 & -34.880 & -8.063  & -20.950 & -5.867 & - \\
% \bottomrule
% \end{tabular}
% }
% \caption{TruthfulQA performance across models (higher is better). The Sig column indicates statistical significance between KARMA\textsubscript{Toxic}  and KARMA\textsubscript{Benign} (Wilcoxon signed-rank test, $\alpha = 0.01$).}
% \label{tab:truthfulqa_all_models}
% \end{table}

\begin{table}[ht]
\centering
\resizebox{\linewidth}{!}{
\begin{tabular}{lcccccc}
\toprule
\textbf{Model} & \textbf{Base} & \textbf{\textsc{karma}\textsubscript{Benign}} & \textbf{\textsc{karma}\textsubscript{Toxic}}  \\
\midrule
Falcon-1B   & -23.265 & \textcolor{gray}{-34.629} & \textcolor{gray}{-31.118} \\
Falcon-7B   & -21.577 & \textcolor{gray}{-35.120} & \textcolor{gray}{-38.410} \\
% LLaMA-1B    & -37.390 & -20.187 & -11.122 \\
LLaMA-8B    & -22.990 & \textcolor{gray}{-36.186} & \textcolor{gray}{-30.984} \\
Pythia-70M  & -17.478 & \textcolor{gray}{-19.440} & -5.435  \\
Pythia-1B   & -23.199 & \textcolor{gray}{-37.390} & -5.436  \\
Pythia-6.9B & -26.817 & \textcolor{gray}{-34.880} & -20.950 \\
\bottomrule
\end{tabular}
}
\caption{TruthfulQA performance across models (higher is better). Gray denotes worse than base.}
\label{tab:truthfulqa_all_models}
\end{table}

The results in Table~\ref{tab:truthfulqa_all_models} evaluate model \textbf{truthfulness} using \textbf{TruthfulQA}, where higher (less negative) scores indicate improved factual accuracy. Overall, performance varies substantially across models and training regimes, with \textsc{karma}\textsubscript{Benign} often leading to declines in truthfulness, suggesting that engagement-oriented optimization can negatively impact factual reliability. These results suggest that \textsc{karma} alignment erodes truthfulness, importantly suggesting that social media data may tend to deprioritize accuracy and prefer rewarding misconceptions, which is not mitigated by \textsc{karma}\textsubscript{Benign}. A potential explanation is catastrophic forgetting induced by PPO. To evaluate this possibility, we evaluate MMLU performance with the expectation that catastrophic forgetting should erode performance there as well. 

% \begin{table}[ht]
% \centering
% \small
% \setlength{\tabcolsep}{4pt}
% \resizebox{\linewidth}{!}{
% \begin{tabular}{lcccccc}
% \toprule
% \textbf{Model} & \textbf{Base} & \textbf{KARMA\textsubscript{Benign}} & $\Delta_{Benign}$ & \textbf{KARMA\textsubscript{Toxic}} & $\Delta_{Toxic}$ & \textbf{Sig} \\
% \midrule
% Falcon-1B   & 0.393 & 0.391 & - & 0.369 & -0.024 & B$\uparrow$ \\
% Falcon-7B   & 0.653 & 0.643 & -0.010 & 0.623 & -0.030 & B$\uparrow$ \\
% LLaMA-1B    & 0.229 & 0.229 &  - & 0.242 & +0.013 & T$\uparrow$ \\
% LLaMA-8B    & 0.541 & 0.537 & - & 0.532 & -0.009 & – \\
% Pythia-70M  & 0.198 & 0.192 & - & 0.195 & - & – \\
% Pythia-1B   & 0.233 & 0.229 & - & 0.229 & - & – \\
% Pythia-6.9B & 0.248 & 0.241 & - & 0.243 & - & – \\
% \bottomrule
% \end{tabular}
% }
% \caption{MMLU performance across models (higher is better). The Sig column indicates statistical significance between KARMA\textsubscript{Toxic}  and KARMA\textsubscript{Benign} (Wilcoxon signed-rank test, $\alpha = 0.01$). $\Delta$ columns with dashes show that the difference from base is not significant.}
% \label{tab:mmlu_all_models}
% \end{table}

\begin{table}[ht]
\centering
\small
\setlength{\tabcolsep}{4pt}
\resizebox{\linewidth}{!}{
\begin{tabular}{lcccccc}
\toprule
\textbf{Model} & \textbf{Base} & \textbf{\textsc{karma}\textsubscript{Benign}} & \textbf{\textsc{karma}\textsubscript{Toxic}} \\
\midrule
Falcon-1B   & 0.393 & \textcolor{gray}{0.391} & \textcolor{gray}{0.369} \\
Falcon-7B   & 0.653 & \textcolor{gray}{0.643} & \textcolor{gray}{0.623} \\
% LLaMA-1B    & 0.229 & \textcolor{gray}{0.229} & 0.242 \\
LLaMA-8B    & 0.541 & \textcolor{gray}{0.537} & \textcolor{gray}{0.532} \\
Pythia-70M  & 0.198 & \textcolor{gray}{0.192} & \textcolor{gray}{0.195} \\
Pythia-1B   & 0.233 & \textcolor{gray}{0.229} & \textcolor{gray}{0.229} \\
Pythia-6.9B & 0.248 & \textcolor{gray}{0.241} & \textcolor{gray}{0.243} \\
\bottomrule
\end{tabular}
}
\caption{MMLU performance across models (higher is better). Gray indicates worse than base.}
\label{tab:mmlu_all_models}
\end{table}

The results in Table~\ref{tab:mmlu_all_models} evaluate core capability and factual recall using \textbf{MMLU}, where higher scores indicate better performance. Overall, performance is consistently degraded by \textsc{karma}, though the degradation is very small. Differences between training regimes are also small in magnitude and generally not statistically meaningful, with most changes on the order of $10^{-3}$ to $10^{-2}$. These results suggest that \textsc{karma} induces only a minor decrease in MMLU performance, indicating that losses in truthful QA are unlikely to be induced by catastrophic forgetting.

\subsection{Results Discussion}

\textsc{karma}\textsubscript{Benign} avoids directly exposing the downstream model to social media data and shows improvements in pragmatics-mediated scenarios, mitigates most biases, and reduces toxicity. In contrast, \textsc{karma}\textsubscript{Toxic} exposes the downstream model to the social media data which is expected to hold toxic content and fails to consistently improve the model, mitigate bias effectively, or reduce toxicity. 

Both methods showed an important side effect, social media reward degraded factual capability in both associated tests. This is the most consistent negative side effect of the approach. Critically, this effect persists under \textsc{karma}\textsubscript{Benign}, where the model has no direct exposure to Reddit data, indicating that the tension between pragmatic reward and factual accuracy is embedded in the reward signal itself rather than introduced by noisy training data. This stands as a critique of social media, suggesting that to be rewarding across communities one typically forgoes strict factuality. Reduced factuality could have significant societal impacts.

LLaMA 8B lies outside the typical behavior of the other models when evaluated on sentiment analysis, toxicity, and sycophancy. We hypothesize that this may be due to the reward model being based on a LLaMA family model which may permit easier intra-family reward hacking in some applications.

\section{Impact of Reward Model Inputs}

To better understand the role of reward model conditioning, we construct a variant in which the reward model is explicitly trained with additional Reddit-specific structural features, including subreddit identity and timestamp metadata (date and time of posting). These features provide contextual signals about conversational norms, audience expectations, and temporal engagement patterns, expected to allow the reward model to more accurately predict interaction outcomes within the Reddit domain. This conditioned reward model is then used to perform PPO-based alignment which we denote as \textsc{karma}\textsubscript{Eroded}. 

On held-out Reddit evaluation data, the conditioned reward model demonstrates improved predictive performance, achieving an AUC of 0.772 and an F1 score of 0.7521, with accuracy, F1 score, and AUC all increasing by approximately 3--4\% relative to the \textsc{karma} reward model trained without such metadata. This indicates that incorporating structured platform-specific context meaningfully enhances the reward model's ability to capture engagement dynamics within its training distribution.

\begin{table}[h]
\centering
\scriptsize
\setlength{\tabcolsep}{4.5pt}
\resizebox{\linewidth}{!}{
\begin{tabular}{llccc}
\toprule
\textbf{Model} & \textbf{Metric} & \textbf{Base} & \textbf{\textsc{karma}} & \textbf{\textsc{karma}\textsubscript{Eroded}} \\
\midrule
\multirow{4}{*}{Falcon 1B}
 & ColBERT Humor    & 0.498  & \textbf{0.514}  & \textcolor{gray}{0.491} \\
 & CrowS-Pairs Bias & -0.133 & \textbf{-0.239} & -0.226 \\
 & RealToxicity     & 0.058  & \textbf{0.041}  & \textcolor{gray}{0.100} \\
 & Sycophancy       & 0.482  & \textbf{0.383}  & \textcolor{gray}{0.593} \\
\midrule
\multirow{4}{*}{Pythia 1B}
 & ColBERT Humor    & 0.500  & \textbf{0.501}  & \textcolor{gray}{0.499} \\
 & CrowS-Pairs Bias & -0.200 & \textbf{-0.431} & -0.300 \\
 & RealToxicity     & 0.071  & 0.053           & \textbf{-0.018} \\
 & Sycophancy       & -0.002 & -0.012          & \textbf{-0.015} \\
\bottomrule
\end{tabular}
}
\caption{Falcon 1B and Pythia 1B performance across Base, \textsc{karma}, and 
\textsc{\textsc{karma}}\textsubscript{Eroded} alignment conditions. 
\textcolor{gray}{Gray} denotes worse than base performance. 
\textsc{karma} improves all metrics over base for both models.}
\label{tab:eroded_comparison}
\end{table}

Interestingly, when deployed as a general-purpose reward signal during \textsc{karma} training, this added information erodes downstream benefits. As a result, the \textsc{karma}-aligned Falcon model aligned using this conditioned reward demonstrates degraded performance and increased variance in behavioral metrics. 
% These findings highlight a tradeoff between reward model fidelity and generalization. While domain-specific conditioning improves in-distribution reward accuracy, it does not produce downstream improvement. 

This result is consistent with \textit{shortcut learning}~\cite{geirhos2020shortcut}, wherein a model achieves strong in-distribution performance by exploiting surface-level features rather than learning the underlying target concept. Subreddit identity and timestamp metadata are highly predictive of karma within the Reddit domain, but carry no generalizable pragmatic signal. A reward model with access to these features can largely bypass pragmatic reasoning entirely, achieving superior predictive accuracy while learning a fundamentally different and less transferable function. 
% The observed downstream degradation under \textsc{KARMA}\textsubscript{Eroded} is therefore a predictable consequence: PPO optimizes toward whatever signal the reward model has internalized, which in this case reflects platform-specific pattern matching rather than conversational appropriateness. 

This finding suggests that reward model evaluation metrics such as F1 and AUC are insufficient proxies for downstream alignment quality, and that careful control over reward model inputs is essential to preserving the generalizability of the learned signal.

\section{Conclusion}

We introduced \textsc{karma} (Karma-Aligned Reward Model Adaptation), a framework for improving context-sensitive conversational behavior in language models through reward learning from  large-scale social interaction data. \textsc{karma} trains a reward model on Reddit conversations to capture implicit pragmatic signals, then applies PPO-based reinforcement learning to align downstream models toward these preferences without requiring direct exposure to social media data during fine-tuning.

Our experiments across multiple model families and scales yield three principal findings. First, \textsc{karma} consistently improves performance on pragmatics-mediated tasks, particularly humor recognition and sentiment classification, supporting our hypothesis that Reddit engagement signals encode meaningful information about conversational appropriateness. These gains hold across model sizes and architectures, with \textsc{karma}\textsubscript{Benign} providing the most consistent improvements. Second, the data distribution used during the PPO training plays a critical role in determining both the benefits and risks of alignment. \textsc{karma}\textsubscript{Benign} optimizes on general benign chat data using the Reddit-trained reward model, reliably reducing toxicity, stereotypical bias, and sycophancy while causing minimal degradation to core factual capability. \textsc{karma}\textsubscript{Toxic}, by contrast, can amplify undesirable behaviors through direct exposure to Reddit's adversarial and inflammatory content, underscoring the value of reward-mediated adaptation as a safer alternative. Reward model predictive accuracy does not translate straightforwardly to downstream alignment quality. A conditioned reward model incorporating subreddit identity and timestamp metadata achieved meaningfully higher F1 and AUC on held-out Reddit data, yet produced degraded and more variable downstream behavior. This suggests that platform-specific heuristics can dominate the more generalizable pragmatic signal, and that reward model design requires careful consideration beyond standard predictive metrics.

One exception worth noting concerns LLaMA 8B, which shows less consistent gains across evaluations. Because the \textsc{karma} reward model is itself a fine-tuned LLaMA 1B model, results for LLaMA-family downstream models may reflect reward model familiarity effects rather than genuine alignment signal, and should be interpreted with caution.

Taken together, these findings suggest a broader principle: social interaction data contains rich and underexplored supervisory signal for pragmatic alignment, but realizing its benefits requires deliberate choices about what information the reward model can access and what data the policy model is exposed to during optimization. \textsc{karma} is a step toward models that better read the room. 

% Future work should extend this framework to richer evaluation of political bias, long-form consistency, and calibration, and explore whether similar reward learning approaches generalize to other large-scale interaction platforms beyond Reddit.

\section{Limitations}

% A key limitation of using Reddit-derived engagement signals is that they serve as a noisy and context-dependent proxy for pragmatic conversational ability like reading the room. While karma and interaction patterns provide scalable supervision and capture rich information about real-world language use, they are shaped by platform-specific norms and may embed demographic skews, implicit biases, and preferences for emotionally salient or humorous content over factual correctness. Although explicit community identifiers are removed, residual structure in the data may still influence the learned reward function and downstream policy optimization. Additionally, engagement signals often conflate multiple correlated factors such as humor, relevance, persuasion, toxicity, and bias, making it difficult to disentangle the source of observed improvements or degradations in model behavior. Another limitation is the scope of evaluation, which does not fully cover dimensions such as political bias, calibration, robustness, or long-form consistency, leaving potential behavioral effects unmeasured. Despite these limitations, the use of structured conversational context and the removal of explicit platform features encourages more generalizable learning, though some degree of domain specificity may still remain.

\textsc{karma} demonstrates consistent improvements in pragmatic alignment across multiple model families and scales, while revealing important constraints that motivate future work. Reddit karma provides a uniquely scalable and naturalistic source of implicit human preference signal, capturing real-world conversational dynamics across 14k communities without the cost or bottlenecks of manual annotation. However, these signals are shaped by platform-specific norms and may embed demographic skews or preferences for emotionally salient content over factual accuracy. We mitigate this through deliberate design choices, removing explicit community identifiers and standardizing inputs into a chat-like format, which our results confirm encourages more generalizable learning, as evidenced by consistent improvements on out-of-distribution benchmarks.

The current evaluation, while broad, does not cover political bias, long-form consistency, calibration, or robustness to adversarial prompting. These dimensions represent natural extensions of the \textsc{karma} framework and are left for future work.

Results for LLaMA-family downstream models should be interpreted with caution. The \textsc{karma} reward model is itself a fine-tuned LLaMA 1B, which may introduce reward model familiarity effects that confound alignment signal for same-family downstream models. LLaMA 1B was excluded from experiments on these grounds, and anomalous LLaMA 8B behavior observed in sentiment analysis, toxicity, and sycophancy evaluations is consistent with this explanation.

%\nocite{*}
\bibliography{custom}

\end{document}